\documentclass[letterpaper, 10 pt, conference]{ieeeconf}  %

\IEEEoverridecommandlockouts                              %

\usepackage[T1]{fontenc}
\usepackage[utf8]{inputenc}

\usepackage{amsmath} %
\usepackage{amssymb}  %
\usepackage{hyperref}
\usepackage{algorithm2e}
\usepackage{graphicx}
\usepackage{url}
\usepackage{bm}
\usepackage{multirow}
\usepackage{cleveref}
\usepackage{eso-pic}
\usepackage[style=ieee,]{biblatex}
\DeclareMathOperator{\atantwo}{atan2}

\usepackage{ifthen}

\makeatletter
\newcounter{IEEE@bibentries}
\renewcommand\IEEEtriggeratref[1]{%
  \renewbibmacro{finentry}{%
    \stepcounter{IEEE@bibentries}%
    \ifthenelse{\equal{\value{IEEE@bibentries}}{#1}}
    {\finentry\@IEEEtriggercmd}
    {\finentry}%
  }%
}
\makeatother

\title{\LARGE \bf
Trajectory Tracking via Multiscale Continuous Attractor Networks
}

\author{Therese Joseph\hskip5em Tobias Fischer\hskip5em Michael Milford%
\thanks{The authors are with the QUT Centre for Robotics, School of Electrical Engineering and Robotics,  Queensland University of Technology, Brisbane, QLD 4000, Australia. Email: {\tt\footnotesize t2.joseph@hdr.qut.edu.au}}%
\thanks{This work received funding from ARC Laureate Fellowship FL210100156 to MM, and from a grant from Intel Labs to TF and MM. The authors acknowledge continued support from the Queensland University of Technology (QUT) through the Centre for Robotics.}%
}

\begin{document}

\maketitle
\thispagestyle{empty}
\pagestyle{empty}

\AddToShipoutPicture*{%
     \AtTextUpperLeft{%
         \put(0,20){
           \begin{minipage}{\textwidth}
              \footnotesize
              Preprint version; final version available at \url{http://ieeexplore.ieee.org}\\
              IEEE/RSJ International Conference on Intelligent Robots and Systems (IROS 2023)\\
              Published by: IEEE
           \end{minipage}}%
     }%
}

\begin{abstract}

Animals and insects showcase remarkably robust and adept navigational abilities, up to literally circumnavigating the globe. Primary progress in robotics inspired by these natural systems has occurred in two areas: highly theoretical computational neuroscience models, and handcrafted systems like RatSLAM and NeuroSLAM. In this research, we present work bridging the gap between the two, in the form of Multiscale Continuous Attractor Networks (MCAN), that combine the multiscale parallel spatial neural networks of the previous theoretical models with the real-world robustness of the robot-targeted systems, to enable trajectory tracking over large velocity ranges. To overcome the limitations of the reliance of previous systems on hand-tuned parameters, we present a genetic algorithm-based approach for automated tuning of these networks, substantially improving their usability. To provide challenging navigational scale ranges, we open source a flexible city-scale navigation simulator that adapts to any street network, enabling high throughput experimentation\footnote{\label{GitHubRepo} \url{https://github.com/theresejoseph/Trajectory_Tracking_via_MCAN/}}. In extensive experiments using the city-scale navigation environment and Kitti, we show that the system is capable of stable dead reckoning over a wide range of velocities and environmental scales, where a single-scale approach fails.

\end{abstract}

\section{Introduction}

Robotic navigation systems encounter several challenges: creating efficient maps, adapting to significant environmental changes, and long-term tracking. In contrast, biological systems have evolved efficient strategies for lifelong navigation while performing tasks such as foraging, migration, and homing. Neuroscience research on spatial representation has identified some of these mechanisms that encode and integrate sensory information to build spatial maps using models such as Continuous Attractor Networks (CAN)~\cite{seung1997learning,arnold1991learning,burak2009accurate}. Robotics has drawn inspiration from this field and developed algorithms such as RatSLAM~\cite{milford2010persistent} and NeuroSLAM~\cite{yu2019neuroslam} that attempt to emulate some of the biological mechanisms of spatial navigation. These systems still face significant challenges in terms of one or more of robustness, scalability, ease of deployment, and adaptability.

In particular, robust navigation systems should be able to operate over a wide range of spatial scales without incurring excessive memory usage. While most CAN navigation systems use single-scale networks for 2D and 3D spaces, a multiscale network can switch modes of operation according to the input scale, making them suitable for long-term navigation.

\begin{figure}[t]
     \includegraphics[width=0.9\linewidth]
      {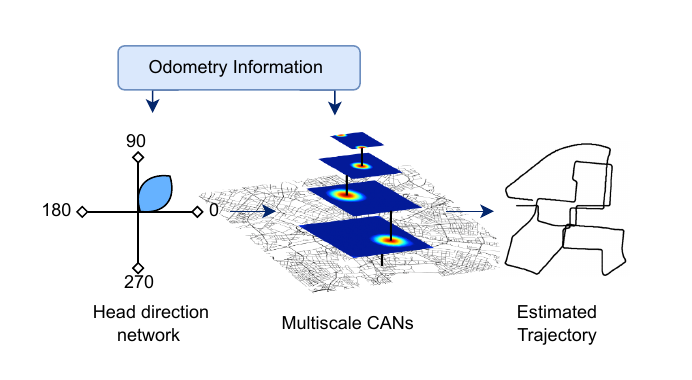}
      \vspace*{-0.2cm}
      \caption{Overview of the Multiscale Continuous Attractor Network (MCAN) architecture: The head direction network on the left processes angular velocity to estimate the robot's heading direction, which is then fed into MCAN along with linear velocity. The MCAN integrates these inputs at multiple scales and generates trajectory estimates -- an example of a generated trajectory is shown on the right. Note that MCAN only uses odometry information.}
    \label{fig:front}
    \vspace*{-0.2cm}
\end{figure}

This paper proposes a new multiscale biologically inspired network, with the following contributions: 
\begin{enumerate}
    \item \textbf{Multiscale Continuous Attractor Networks (MCAN):} Our proposed Multiscale Continuous Attractor Networks (MCAN) is a bio-inspired neural network architecture with multi-scale parallel spatial neural networks building on previous theoretical models to accurately encode a wide range of velocity inputs and enable large scale trajectory tracking (\Cref{fig:front}).   
    \item \textbf{A tuning method using genetic algorithms:} The reliance of previous systems on hand-tuned parameters is overcome by presenting a genetic algorithm-based approach for automated tuning of the MCAN and a head direction network. This optimization method automatically identifies high performance parameter spaces in large CAN networks. 
    \item \textbf{A city-scale navigation simulator:} To provide challenging navigational scale ranges, we contribute a flexible city-scale navigation simulator that adapts to any street network, enabling high throughput experimentation for evaluating path integration performance. %
\end{enumerate}

The paper is structured as follows: In Section~\ref{sec:relatedworks}, we will provide an overview of the brain's navigational mechanisms, attractor networks, and related research. Section~\ref{sec:methodology} presents the methodology, which explains the process of simulating robot trajectories within city road networks, optimizing network performance using genetic algorithms, and utilizing the multiscale network architecture with attractor dynamics. The results of extensive experiments using both the new simulator and Kitti are presented in Sections~\ref{sec:experimentalsetup} and~\ref{sec:results}, where we quantitatively compare a single-scale baseline to the proposed multiscale system. Finally, Section~\ref{sec:conclusion} concludes the paper with recommendations for future work.

\section{Related Works}
\label{sec:relatedworks}

Section~\ref{subsec:navigationmechanisms} introduces spatial cells in the hippocampus, including place cells, head direction cells, grid cells, and border cells, that play a critical role in mammalian navigation.Section~\ref{subsec:theoreticmodels} discusses the development of several computational models that integrate linear and angular velocity cues for path integration while mimicking the characteristics of spatial cells. Section~\ref{subsec:bioinspiredrobotnav} reviews bio-inspired robotic systems that emulate the behaviour of spatial cells within real environments using robotic hardware. Section~\ref{subsec:tuningcans} then presents methods for tuning CANs using optimization techniques, and finally, Section~\ref{subsec:simsfornavigation} discusses simulators for robot navigation.

\subsection{Navigation Mechanisms in the Brain}
\label{subsec:navigationmechanisms}
Mammalian navigation is a complex process that involves various spatial cells in the hippocampus, such as place cells, head direction cells, and grid cells. Place cells \cite{o1976place} are neurons that fire at unique spatial locations, representing places within an environment. Head direction cells \cite{taube1998head} encode an animal's orientation and provide information about its heading. Grid cells \cite{hafting2005microstructure,moser2014grid,banino2018vector} use a tessellated grid pattern to integrate direction and speed, enabling efficient encoding of large environments \cite{grieves2017representation}. Border cells and object vector cells are additional spatial cells that provide information about the boundaries within an environment along with the distance and direction of objects within it \cite{hoydal2019object, bordercell}. These biological mechanisms offer an alternative to classical robotic Simultaneous Localization And Mapping (SLAM) algorithms by integrating sensory visual and motion cues to update estimates of a location while building a cognitive map of the environment. 

\subsection{Theoretic Computational Neuroscience Models}
\label{subsec:theoreticmodels}
The discovery of spatial cells led to the development of several computational models that can integrate linear and angular velocity cues for path integration while mimicking the characteristics of spatial cells. Attractor networks were proposed to model head direction~\cite{skaggs1994model} using neurons in a ring structure with a stable bump of activity, which shifts based on angular velocity inputs. They were also used to model grid cells with 2D CAN models that combine head direction and speed for shifting activity bumps with toroidal connections at the boundary~\cite{fyhn2004spatial,mcnaughton2006path,burak2009accurate}. %

Early works \cite{zhang1996representation,samsonovich1997path,widloski2014model} implemented attractor models with unsupervised learning, while supervised models \cite{hahnloser2000permitted,arnold1991learning} incorporated learning rules or error signal feedback. Other models \cite{banino2018vector,kanitscheider2017emergence,seung1997learning} have used backpropagation and architecture constraints to form computational navigation models. 

Although these theoretical models cannot fully replicate the complexity and nuance of how the brain solves navigational problems, they offer valuable insights into the computations that the brain employs for integration, error correction, and learning \cite{khona2022attractor}. 

\subsection{Bio-inspired Robot Navigation}
\label{subsec:bioinspiredrobotnav}
The discovery of navigational mechanisms in the brain also resulted in algorithms that emulate the behaviour of spatial cells in physical robotic hardware, demonstrating navigational capabilities through bio-inspiration. Early works include \cite{gaussier2002view,cuperlier2007neurobiologically,milford2010persistent} which rely on place cells along with curated mechanisms like ``view cells'', ``transition cells'' and ``pose cells'' based on standard robotics principles. %
Extensions of these works include NeuroSLAM~\cite{yu2019neuroslam} which extends RatSLAM~\cite{milford2010persistent} to a 3D space and \cite{li2022brain} which corrects absolute heading using polarization of light as seen in Desert Ants. 

Grid cell mechanisms have also been the basis for numerous multiscale systems in robotics such as the large-scale aerial mapping system developed by Hausler et al.~\cite{hausler2020bio}, a grid-cell inspired place recognition system that utilizes homogeneous maps at varying scales \cite{chen2015bio}, and a system for multiscale path integration in 3D for unmanned aerial vehicles (UAVs) \cite{yang2021path}. These systems demonstrate the potential of grid cell-based models for solving complex tasks in different domains.

\subsection{Tuning CANs}
\label{subsec:tuningcans}
Existing methods for tuning CANs have relied on either optimization techniques or hand-crafted network parameters to achieve stable activity that integrates input signals accurately. In early works, DeGris et al.~\cite{degris2004spiking} used a Genetic Algorithm to fine-tune 1D CAN networks in a spiking neuron model. Dall'Osto et al.~\cite{dall2018automatic} proposed an automatic calibration method utilizing global optimization methods, while Menezes et al.~\cite{menezes2020automatic} presented an iterative closest point algorithm for automatic tuning of RatSLAM. More recently, Fox et al.~\cite{fox2022new} proposed a new evolutionary dynamic optimization method for shifting attractor peaks, along with a benchmark suite. However, with the growth of network complexity in scale and dimension, optimizing additional parameters becomes increasingly challenging.

\subsection{Simulators for Robot Navigation}
\label{subsec:simsfornavigation}
Simulations play a crucial role in enabling rapid experimentation and development of any system. OpenStreetMap (OSM) is a crowd-sourced map that includes road network information obtained from portable GPS devices, and it has been used in several systems. For example, Brubaker et al.~\cite{brubaker2013lost} performed self-localization using OSM maps and visual odometry. Fleischmann et al.~\cite{fleischmann2017using} used OSM paths to assess the quality of GNSS signals during navigation. Li et al.~\cite{li2021openstreetmap} combined OSM road network paths with onboard sensors for path tracking. By integrating OSM maps with other data sources, these systems showcase the versatility and potential of OSM in various applications and motivated the simulation developed within this work.

\section {Methodology}
\label{sec:methodology}
Our work aims to enhance the ability of robots to navigate through large-scale environments by introducing a new multiscale extension to continuous attractor models. In Section~\ref{subsec:attractdyna}, we describe the dynamics of the 2D attractor network, which serves as the foundation of our approach. We then present the multiscale network extension in \mbox{Section~\ref{subsec:multiscalenetwork}}, which enables the network to handle large-scale velocity inputs and accurately track trajectories over extended periods. Additionally, we introduce the head direction network in Section~\ref{subsec:hdnetwork}, which processes the robot's heading direction information and feeds it into the multiscale network. To further improve the usability and performance of our approach, we employ a genetic algorithm-based tuning approach, which is detailed in Section~\ref{subsec:tuning}. Our methodology is evaluated in a new city-scale navigation simulator that we introduce in Section~\ref{subsec:simulator}, providing a comprehensive evaluation of our approach's ability to track motion through a range of large, varied street networks.

\subsection{Attractor Dynamics}
\label{subsec:attractdyna}

Attractor networks are widely used in neural modelling \cite{yu2019neuroslam} \cite{milford2010persistent} \cite{khona2022attractor} due to their ability to represent and maintain stable patterns of activity, even in the presence of noise and input variability. In this subsection, we describe the attractor dynamics used in our model, which is based on a 2D grid of neurons with recurrent excitatory connections. Specifically, a 2D attractor network $\mathbf{X}$ is defined as an $N_x\times N_y$ grid of neurons.

\subsubsection{\textbf{Initilization}}
At the beginning of the simulation, the neurons in the attractor network are initialized using a 2D Gaussian function with standard deviation $(\sigma_x, \sigma_y)$, which determines their initial state of activation. We initialize the activity in an $A\times A$ region centred around neuron $(x_0, y_0)$, resulting in the following activation profile:

\begin{equation}
\mathbf{G}(i,j) = \exp\left(-\frac{(i-x_0)^2}{2\sigma_x^2} - \frac{(j-y_0)^2}{2\sigma_y^2}\right)
\end{equation}

\begin{equation}
    \mathbf{X}(0,i,j) =
    \begin{cases}
        \mathbf{G}(i,j), & \text{if } i\in[x_0-A,x_0+A], \\
          & \hspace{-0.4cm}\qquad j\in[y_0-A,y_0+A] \\
        0, & \text{otherwise}
    \end{cases}
\end{equation}

\subsubsection{\textbf{Network update}}
At the end of each time step $\delta t$, the activity of each neuron is divided by the L2 norm of the activity across the entire network. This ensures that the maximum magnitude of the activity vector is always one and prevents explosive activity growth:
\begin{equation}
\mathbf{X}(t+\delta t,i,j) = \frac{\mathbf{X}(t,i,j)+\mathbf{C_f}(i,j)+\boldsymbol\epsilon(i,j)-\mu}{\left\| \mathbf{X}(t,i,j)+\mathbf{C_f}(i,j)+\boldsymbol\epsilon(i,j)-\mu \right\|}
\end{equation}
In this equation, the numerator represents the total input to each neuron, which is the sum of its current activity $\mathbf{X}(t,i,j)$, the activity of neurons shifted into its place field based on the input velocity $\mathbf{C_f}(i,j)$, the excitation caused by nearby active neurons $\boldsymbol\epsilon(i,j)$, and the global inhibition term $\mu$.

\subsubsection{\textbf{Copying and shifting the activity packet}}

After initialization, the attractor network is updated by injecting the activity of active neurons back into the network with an integer offset $\alpha_x$ and $\alpha_y$ corresponding to the input velocity, which shifts the activity packet across the grid. To achieve this, we copy the weights $\mathbf{C}$ of active neurons (i.e., those with an activity greater than zero) into the network, with wraparound connections at the edges. Specifically, we use the following equation:
\begin{equation}
\mathbf{C}_{(i',j')} = \begin{cases}
\mathbf{X}_{(t,i,j)}, & \text{if } \mathbf{C}_{(i',j')} >0, \\
0 & \mathrm{otherwise},
\end{cases}
\end{equation}
where $(i',j')$ is the shifted position of the active neuron $(i,j)$, computed as $(i',j') = \bigl((i+\alpha_x) \bmod N_x,  (j+\alpha_y)\bmod N_y\Bigl)$. 

\subsubsection{\textbf{Fractional shifts}}
In addition to integer shifts, the attractor network can also handle fractional shifts, which allow the activity packet to move more smoothly across the grid. To achieve this, we use linear interpolation to modify the copied weights $\mathbf{C}(i,j)$ with a fractional offset amount $\alpha_f$ and a confidence parameter $\gamma$, resulting in the following fractional copy function:
\begin{align}
\begin{split}
      \mathbf{C_f}(i,j) &= \gamma\big((1 - \alpha_f)\cdot \mathbf{C}(i,j) \\&+ \alpha_f \cdot \mathbf{C}((i+1) \bmod N_x,(j+1)\bmod N_y)\big).
\end{split}
\end{align}

\subsubsection{\textbf{Excitation}}
After the activity packet has been shifted across the grid, the active neurons excite a region of size $E\times E$ (such that $E$ represents the excitation radius), with the strength of the excitation being scaled by the weight, $W_n$, of each neuron. Specifically, we use a 2D Gaussian function to compute the excitation $\epsilon(i,j)$ of neuron $(i,j)$, resulting in the following expression:
\begin{equation}
\boldsymbol\epsilon(i,j)=\sum_{n=1}^{N_x \times N_y}  W_n\cdot\exp\left(-\frac{(i-x_n)^2}{2\sigma_x^2} - \frac{(j-y_n)^2}{2\sigma_y^2}\right).
\end{equation}
As for the initialization, if the neuron index is outside the range $x_n-E < i < x_n+E$ and $y_n-E < j < y_n+E$, then the activity is set to 0.

\subsubsection{\textbf{Inhibition}}
To prevent the network from explosive activity growth, a global inhibition term is applied. The global inhibition $\mu$ that is applied to each neuron is computed by summing the activities of all neurons and scaling the result by an inhibition factor $\phi$:

\begin{equation}
\mu=\sum_{i=1}^{N_x} \sum_{j=1}^{N_y} \mathbf{X}_{i,j}\times \phi.
\end{equation}
This inhibition term acts as a normalizing factor, reducing all neurons' activity when the total network activity becomes too high.

\subsection{Multiscale Attractor Network}
\label{subsec:multiscalenetwork}

The multiscale attractor network builds upon the capabilities of the single-scale attractor network by incorporating multiple 2D networks, each with its own unique scale resolution. This allows the network to capture information at various scales, which is especially relevant in spatial navigation where the agent's velocity can span a large range. By selecting the input network with the closest spatial resolution to the agent's speed, the network can effectively integrate motion information using attractor dynamics and enable accurate position tracking.

\subsubsection{\textbf{Storing network wraparound}}
 Continuous attractor networks in 2D have toroidal manifolds, i.e.~the surface wraps around and reconnects with itself in the $x$ and $y$ dimension, forming a seamless and continuous space that has a torus or donut shape. This enables more efficient computation as neurons on the edge can receive input from the opposite edge. However, the toroidal manifold of the network poses a challenge in terms of preventing decoded position reset when activity wraps around an edge. By storing the distance travelled in wraparound buffers, our algorithm ensures that the agent's position is accurately updated when the activity wraps around an edge, preventing position reset and enabling smooth navigation in the environment. 
 
 For example, if the decoded x position from the network decreases while the agent is facing right (270 < $\theta$ < 90), the wraparound buffer is incremented by the network size to store the magnitude of the position reset. Similarly, if the decoded x position from the network increases while the agent is facing left (90 < $\theta$ < 270), the wraparound buffer is decremented by the network size. In the 2D network, two wraparound buffers are used to store the distance travelled in $x$ and $y$ dimensions.

\subsubsection{\textbf{Position decoding}} 
The final step in the process is to decode the estimated position of the agent using the combined activity of all networks and the wraparound buffer. This is accomplished by taking the circular mean of the most active neuron's row and column, with each network's activity weighed by its corresponding scale resolution. The resulting position estimation is therefore a combination of the information captured at different scales, resulting in a more accurate representation of the agent's true position. Specifically, the decoded position of $M$ networks, each of size $N$ and  scale $s$ is:

\begin{equation}
\small
D= \sum_{j=1}^{M} s(j)\cdot \frac{N}{2\pi} \atantwo\big(\sum_{i=1}^{N}  W_i \sin(i \frac{N}{2\pi}),\sum_{i=1}^{N}  W_i \cos(i\frac{N}{2\pi})\big),
\end{equation}
\normalsize
where $N$ is $N_x$ for decoding the $x$ position and $ N_y$ for decoding the $y$ position, as defined previously.

\subsection{Head Direction Network}
\label{subsec:hdnetwork}
The head direction network plays a crucial role in the navigation system and is also modeled by a continuous attractor network, which is a well-established computational model in the field of neuroscience~\cite{taube1998head,skaggs1994model}. It has a 1D ring structure, and similar to the 2D system, it uses the same tuning parameters ($A, E, \gamma, \phi$) and attractor dynamics. At the beginning of the simulation, the 1D attractor network is initialized with a 1D Gaussian of radius $A$. The activity in the network is updated by shifting activity packets scaled by $\gamma$. Excitation and inhibition are applied with radius $E$ and inhibition factor $\phi$, respectively. The activity in the network is normalized after each update to maintain a stable firing rate.

The network estimates the current heading angle of the agent using the circular mean method. The circular mean is calculated by taking the weighted average of the cosine and sine of the firing angles of the neurons in the network. This method is used to ensure that the direction estimate is robust to circular data, where the values wrap around after reaching the maximum or minimum value. Overall, the head direction network provides an estimate of the agent's heading angle, which is used in conjunction with the multiscale attractor network to update the agent's position.

\subsection{Network Tuning with Genetic Algorithm}
\label{subsec:tuning}
To ensure stable dynamics and accurate velocity integration, the continuous attractor network parameters activation radius $A$, excitation radius $E$, motion confidence $\gamma$, and inhibition factor $\phi$ must be carefully tuned.  
We use a genetic algorithm \cite{katoch2021review} to automate this tuning process. Genetic algorithms are a type of optimization algorithm that is commonly used to explore the fitness landscape of a set of parameters and find the optimal solution. The goal of the tuning process is to ensure that the attractor network exhibits stable dynamics and accurate velocity integration. Each parameter has an operating range that is dependent on the network size and the number of dimensions.

To explore the fitness landscape, the genetic algorithm generates an initial population of potential solutions with varying parameter values. The fitness function is evaluated for each member of the population, and the fittest individuals are selected to be the parents of the next generation. The algorithm then creates three children from the fittest parents by mutating their genes with a mutation probability. This is further detailed in Algorithm~\ref{alg: genetic algorithm}. Fitness is evaluated in parallel at the start of each generation to reduce computing time.

The fitness function used in this case evaluates the performance of the attractor network in integrating velocity information and providing accurate estimates of position and heading. The velocity is sampled from a uniform distribution within a desired operating range to avoid overfitting to a specific input trajectory.

\begin{algorithm}[t]
\vspace{2mm}
\KwData{population $P$, population size $N$, fitness function $f$, mutation rate $r_m$, gene ranges $\rho$, mutation amount $\mu$}
\KwResult {Fittest Child Genome [$A, E, \gamma, \phi$]}
Generate initial population;\\
\While{generation $<$ max generation}{
    $fitnesses \gets$ $f(P_i)$ $\forall i \in P$;\\
    $parents \gets $ top 25\% of $P$ ordered by $fitnesses$;\\
    Clone each parent into three $children$;\\
    \For {g $\in$ $children$} {
        \If{ $U(0,1) < r_m$ }{
        g $\gets$ [g(i)+$N(0,\mu)$,  $ \forall i \in [1,c.size]$];\\
        repeat until c is within $\rho$;\\}}
}
\caption{Tuning Attractor Networks parameters activation radius $A$, excitation radius $E$, motion confidence $\gamma$, and inhibition factor $\phi$ with a Genetic Algorithm}
\label{alg: genetic algorithm}
\end{algorithm}

Overall, the genetic algorithm provides an automated way to fine-tune the parameters of the continuous attractor network, which is essential for the system to work accurately in a variety of environments.

\subsection{City Scale Navigation Simulator}
\label{subsec:simulator}
Our City Scale Navigation (CSN) simulator is a tool used to evaluate the performance of our proposed system in realistic scenarios. The simulator generates trajectories based on real-world road networks within a 10km $\times$ 10km region from major cities. This enables the evaluation of the system's ability to navigate through complex and dynamic urban environments. The simulator extracts road network data from Open Street Maps to generate an occupancy map consisting of traversable, realistic roads. The occupied cells in the map represent the drivable areas of the road network. A path-finding distance transform algorithm \cite{corke2021not}, is then used to find the optimal route between two randomly generated points on the road map.

Once the sample paths are generated, they can be traversed using the kinematics of a bicycle motion model \cite{BicycleModel}, which is a common model used in the navigation of ground vehicles. During the traversal of the paths, motion information such as linear and angular velocities are recorded. This motion data is then used to evaluate the performance of the system, such as the accuracy of the estimated position and heading, the stability of the continuous attractor network, and the effectiveness of the buffer to prevent position resetting.

\section{Experimental Setup}
\label{sec:experimentalsetup}
\subsection{Implementation Details}
The MCAN trajectory tracking system was implemented in Python3 using standard libraries. The head direction network consists of 360 neurons that were tuned using a genetic algorithm to accurately integrate angular velocity inputs to produce an estimate of agent orientation. The 2D networks in the multiscale system was generated with 100$\times$100 neurons and 4 scales with scale ratios incremented by a factor of 4, i.e., $(0.25, 1, 4, 16)$. This was suitable for the desired operating range of 0-20 m/s (0-72 km/h). For a fair comparison, the single-scale CAN was implemented with 200x200 neurons, so both systems had a total of 40000 neurons.  

The CANs were tuned using 24 genomes mutated and evaluated for 20 generations with 14 parallel processes for fitness evaluation. Based on the network size, the parameter ranges were set to $A \in [1, 10]$, $E \in [1, 10]$, $\gamma \in [0, 1]$, and $\phi \in [0.00001, 0.005]$.

\subsection{Datasets}
The performance of our system was evaluated on the City scale navigation simulator (Section~\ref{subsec:simulator}) and the Kitti Odometry dataset~\cite{geiger2012we}, which is commonly used to benchmark robotic systems. The navigation simulator was used to generate trajectories based on road networks within a 10 km $\times$ 10 km region from Tokyo, Berlin, Brisbane, and New York. These road maps were converted into occupancy grids, and a path-finding distance transform algorithm from the Robotics Toolbox for Python \cite{corke2021not} was used to find an optimal route between two points on the road map. Our city scale simulator was used to generate paths through Tokyo,  Brisbane, Berlin and New York, covering distances of 38 km, 70 km, 167 km, and 63 km, respectively. Using both the simulator and the Kitti Odometry dataset, we evaluated the proposed system in a variety of realistic scenarios.

\subsection{Evaluation Metrics}
Our networks were evaluated using the Absolute Trajectory Error (ATE), which is a common metric used in trajectory estimation systems~\cite{zhang2018tutorial}. ATE is defined as the Root Mean Square Error (RMSE) between the estimated and desired trajectory after alignment. In order to ensure a fair comparison between different datasets, ATE was averaged over the total distance, resulting in ATE/meter. While we used ATE to evaluate the performance of our networks, we utilized the Sum of Absolute Differences (SAD) method~\cite{milford2012seqslam} during the tuning process to measure the fitness of the networks.

\section{Results and Discussion}
\label{sec:results}
This section provides a comparison of single-scale CAN versus the proposed multiscale system.  We evaluate the performance of our system on multiple trajectories generated from simulation and existing datasets in Section~\ref{subsec:Performance}. This is followed by an analysis of the tuning mechanism in Section~\ref{subsec:GA}.

\begin{table}[t]
\vspace{2mm}
\caption{Comparison of ATE per meter}
\label{tab:atecomparison}
\renewcommand{\arraystretch}{1.2}
\setlength{\tabcolsep}{2.5pt}
\label{T_results}
\centering
\begin{tabular}{l|cc}

Dataset & Single-scale & Multiscale\\\hline
 Tokyo (CSN simulation)& 1.093 $\pm$ 0.110 & \textbf{0.068 $\pm$ 0.010}\\

 New York (CSN simulation)& 0.893 $\pm$ 0.137& \textbf{0.070 $\pm$ 0.028}\\

 Brisbane (CSN simulation)& 0.934 $\pm$ 0.102 & \textbf{0.070 $\pm$ 0.019}\\

 Berlin (CSN simulation)& 0.896 $\pm$ 0.166& \textbf{0.046 $\pm$ 0.021}\\

 Kitti (odometry)& 0.136 $\pm$ 0.138 & \textbf{0.041 $\pm$ 0.026}\\
\end{tabular}
\vspace{-0.2cm}
\end{table}

\begin{figure}[t]
\vspace{2mm}
     \centering
     \includegraphics[width=0.99\linewidth]{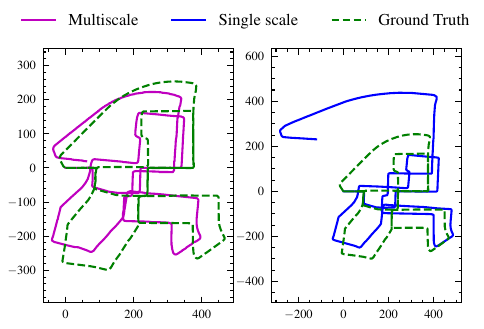}  
    \caption{Comparison of trajectory tracking (in meters) between single-scale and multiscale CAN on the Kitti dataset plotted against ground truth, where the single-scale CAN performance degrades as velocity increases.}
    \label{fig:5.6}
\end{figure}

\begin{figure}[t]
     \centering
     \includegraphics[width=0.8\linewidth]{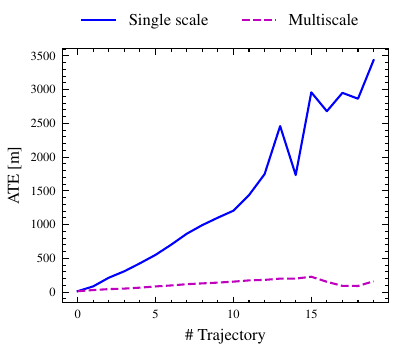} 
    \caption{A Comparison between multiscale and single-scale attractor networks models on Berlin City Scale Navigation Simulation dataset. The two networks are tested on 20 trajectories with increasing velocity ranges to evaluate the invariance to velocity within the multiscale network.}
    \label{fig:5.2}
\end{figure}

\begin{figure}[t]
\vspace{3.7mm}
    \centering
    \includegraphics[width=0.99\linewidth]
    {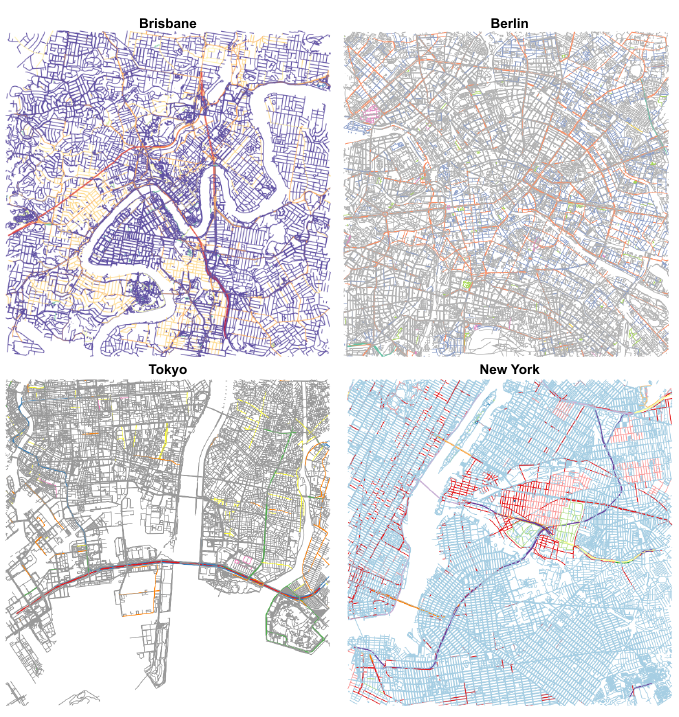}  
    \caption{Maps generated from the City Scale Navigation Simulation used to generate test trajectories: Brisbane, Berlin, Tokyo and New York, with colours representing the road speeds in each city. For example, the highways in Brisbane are shown in orange while the residential roads are in purple.}
    \label{fig:5.1}
\end{figure}

\begin{figure}[t]
\vspace{2mm}
     \centering
     \includegraphics[width=0.99\linewidth]{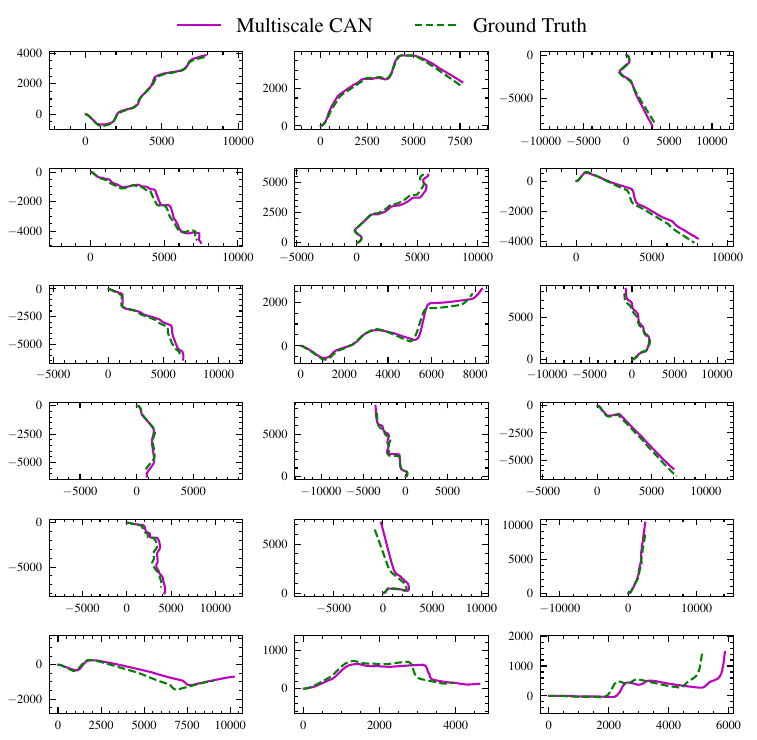} 
    \caption{Trajectory tracking with multiscale networks across 18 tracks from the Berlin City Scale Navigation Simulation Dataset with a total distance of 167km. The tracks, sorted based on tracking error, show varied and complex trajectories that emulate Berlin road networks.}
    \label{fig:5.5}
\end{figure}

\begin{figure}[t]
     \centering     \includegraphics[width=0.85\linewidth]{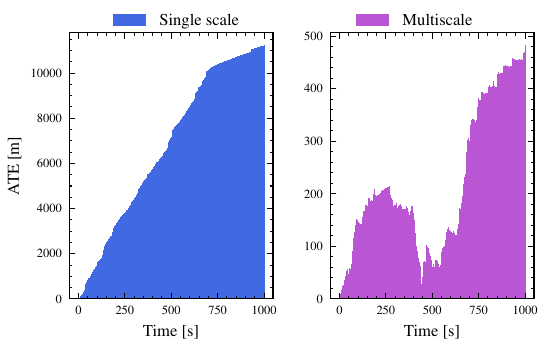}
    \caption{ATE error over time for single-scale and multiscale networks integrating velocities ranging from 0-20m/s for a single trajectory from the Berlin City Scale Navigation Simulation dataset. Note the scale difference between plots.}
    \label{fig:5.7}
\end{figure}

\begin{figure}[t]   
\vspace{2mm}
    \centering
    \includegraphics[width=0.8\linewidth]{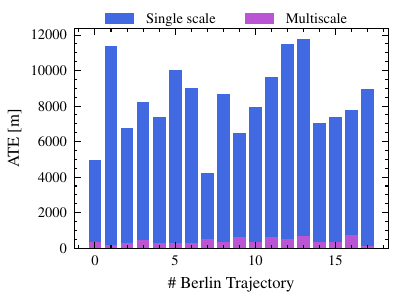}  
    \caption{Average Trajectory Error (ATE) within local segments of simulated trajectories through Berlin. The trajectories are realigned after each segment so the error doesn't accumulate across trajectories. The multiscale error ranges from 0-500m; whereas, the single-scale ranges from 0-11000m.}
    \label{fig:5.8}
\end{figure}

\begin{figure}[t]
     \centering
     \includegraphics[width=0.85\linewidth]{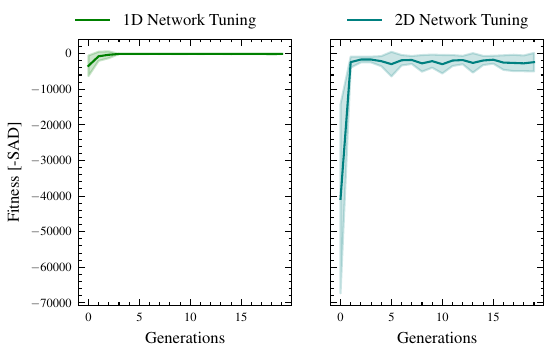}  
    \caption{Fitness evolution of the head direction network [left] and Multiscale Network [right] over 20 generations. Here the fitness is displayed as -SAD which approaches 0 over generations, as the algorithm converges.}
    \label{fig:5.3}
\end{figure}

\subsection{Comparison of Multiscale versus Single-Scale}
\label{subsec:Performance}

Our experiments demonstrate that incorporating a multiscale system can lead to significant performance enhancements on the standard CAN model. Table~\ref{tab:atecomparison} presents the average translational error (ATE) for the single-scale CAN and our proposed MCAN model on five different datasets. We observed that the ATE for MCAN was orders of magnitude lower than that of the single-scale baseline, with improvements evident across all tasks. The City Scale Navigation simulations showed the most significant difference between MCAN and CAN, likely because of their more extensive range of input velocities. On the Kitti dataset, MCAN showed performance improvements specifically in areas of the path with higher velocities and minor improvements in the other regions, as depicted in Figure~\ref{fig:5.6}. 

Furthermore, we evaluated the impact of the input velocity range on the performance of the single-scale and multiscale systems. Figure~\ref{fig:5.2} shows the ATE for different velocity ranges for the OSM simulation dataset. We can observe that the ATE of the single-scale network increases linearly with the velocity range, whereas the ATE of the MCAN increases with a smaller gradient. This demonstrates that MCAN can handle varying velocity ranges and maintain high accuracy, even when faced with challenging trajectories, generated from the tracks seen in Figure~\ref{fig:5.1}.

Figure~\ref{fig:5.5} provides further insights into the behaviour of MCAN. It shows track segments from the Berlin CSN simulator with increasing tracking errors. Trajectories with multiple 90-degree turns have increased errors in comparison to trajectories with smaller turns.

However, these errors are orders of magnitude less than the single-scale model. An example of this is in Figure~\ref{fig:5.7}, where the error accumulates up to 10000 meters within the single-scale run and MCAN has a maximum error of 450 meters with less accumulation of error over time. This is further supported by Figure~\ref{fig:5.8} where the MCAN error is consistently lower in all 18 tracks through Berlin.

\subsection{Tuning Performance}
\label{subsec:GA}

The genetic algorithm tuning procedure was successful in achieving good performance despite the added complexity of multiple interacting models. Both heading direction (1D) and multiscale (2D) tuning converged after about 3 generations, as shown in Figure~\ref{fig:5.3}. The multiscale tuning had limited fitness improvements after generation 2, suggesting that either there is an upper limit on what fitness is realistically possible or that the algorithm is prone to getting stuck in local optima.

\section{Conclusions and Future Work}
\label{sec:conclusion}
In conclusion, our proposed multiscale architecture, coupled with the genetic algorithm tuning procedure, provides a step forward in making bio-inspired systems work with both large scale simulated and real-world data and capable of handling large velocity ranges. This is a key step towards making bio-inspired networks competitive with conventional robotic navigation systems.

Through the development of a multiscale cognitive architecture, we were able to significantly enhance the performance of the continuous attractor model. Our approach results in a system capable of handling a wide range of velocities and complex environments. The proposed genetic algorithm tuning procedure allowed for efficient and effective optimization of network parameters, reducing the need for manual tuning and increasing the scalability of the system. Our results show that the genetic algorithm can successfully optimize both the heading direction and multiscale parameters, leading to improved performance in navigation tasks.

While our proposed system has shown promise, there is potential for future work to address some existing limitations. One limitation is the absence of landmarks or memorized locations of previously visited areas, which means errors could eventually accumulate without a corrective mechanism. Integrating a loop closure component will move this from a dead reckoning system to a full mapping and localization system. Although we conducted extensive simulation experiments drawing upon real world city networks as well as real-world data from Kitti, further live experimentation on deployed robot platforms will provide insights that can drive and inform future model development.

\printbibliography

\end{document}